\documentclass[runningheads]{llncs}

\usepackage{amssymb}
\usepackage{lineno,booktabs}
\usepackage[ruled,vlined]{algorithm2e}
\usepackage[table]{xcolor}
\usepackage{stackengine}
\usepackage{graphicx,subfigure}
\usepackage{amsmath,amssymb}
\usepackage{hyperref}
\usepackage{amsmath}
\usepackage{flushend}

\begin{document}

\title{ A new Local Radon Descriptor for Content-Based Image Search}

\author{Morteza Babaie \orcidID{0000-0002-6916-5941} \and
Hany Kashani \and
Meghana D. Kumar \and
H.R. Tizhoosh \orcidID{0000-0001-5488-601X}}
\authorrunning{To appear in International Conference on AI in Medicine (AIME 2020)}
%
\institute{Kimia Lab, University of Waterloo, Ontario, Canada \\ \email{mbabaie,kashani,meghana,tizhoosh@uwaterloo.ca}}

\maketitle

\begin{abstract}
Content-based image retrieval (CBIR) is an essential part of computer vision research, especially in medical expert systems. Having a discriminative image descriptor with the least number of parameters for tuning is desirable in CBIR systems. In this paper, we introduce a new simple descriptor based on the histogram of local Radon projections. We also propose a very fast convolution-based local Radon estimator to overcome the slow process of Radon projections. We performed our experiments using pathology images (KimiaPath24) and lung CT patches and  test our proposed solution for medical image processing. We achieved superior results compared with other histogram-based descriptors such as LBP and HoG as well as some pre-trained CNNs.

\end{abstract}
\section{Introduction}
Over the past decades, there has been a dramatic increase in capturing and storing data in the form of digital images. In medical imaging, for example, the volume of stored data is expected to exceed more than $2,314$ Exabytes ($10^9$ GB) by 2020  which is an exponential growth from $153$ Exabytes in 2013 \cite{kejariwal2015real}. Demands to extract information from these massive archives is growing day to day. Content-based image retrieval (CBIR) system undoubtedly is considered a necessary way to extract this information. CBIR has long been a subject of great interest in a wide range of fields, from searching pictures of  celebrities on the Internet to helping radiologists and pathologists to  make a more accurate diagnosis by providing them access to the images of similar cases. In general, CBIR is referred to searching a dataset to retrieve similar images to a query image \cite{rui1997image}. In the medical community, content-based medical image retrieval (CBMIR) refers to the same tasks in the medical image domain. However, in CBMIR the semantic gap between algorithms and the experts in capturing the similarity is more crucial  \cite{alzu2015semantic}. In addition, the size and quantity of medical images are quite overwhelming.  A desirable searching method is expected to return  similar images for any query image in a reasonable time. To this end, we used a practical and well-known technique in the medical domain, called Radon transform, to describe the visual features in medical images. We utilized Local Radon Projections (LRP) in a very fast and accurate way. Local Radon Descriptor (LRD) has been applied to medical images with large window size and a relatively high number of parameters to tune \cite{babaie2017local}. However,  implementations of Radon transform on the large patches are computationally slow because it interpolates the pixel values from multiple pixels. In this work, we estimated the Radon transform in small patches by convolving a series of designed filters in the whole image and then reading small patches to achieve a fast method to apply in CBMIR. Also, the presented version of LRP is quite simple and no parameter to tune.  

The rest of this paper is organized as follows: The next section explores related works in CBMIR and descriptors fields as well as the literature related to the Radon Transform and filter-based image processing. The methodology section describes details of the LRP and its fast implementation. And finally, the results of our experiment are presented at the end of the paper followed by conclusions.

\section{Related Works}
Artificial Intelligence (AI) can play an important role in the field of medical image analysis. In recent years, there has been an increasing interest in applying AI in the medical domain to help clinicians with the diagnosis and treatment of the diseases  \cite{bankman2008handbook}. Detection, classification, segmentation, and retrieval are the mainstream fields that are currently subject to research in various domains of medical imaging \cite{de2016machine}. All the above-mentioned approaches are mainly designed to work on a specific organ and modality (e.g., analysis methods on brain images produced by MRI \cite{bauer2013survey} or lung segmentation in X-ray images). However, these methods generally need carefully labeled medical image data which is not a feasible task, due to the expense related to the specialist's involvement and the huge size of datasets \cite{greenspan2016guest}. CBIR, unlike all other above-mentioned methods, has the ability to work efficiently by a small amount of labeled data and even raw data in large datasets \cite{8253476}. However, labeled data can help to overcome the semantic gap (perhaps the greatest challenge in CBIR) between algorithms and expert regarding perceiving the image similarity \cite{ghosh2011review}.  
 
  Historically, early research in the field of image search focused on text-based systems that return similar images within the same anatomical region, with the same orientation and using the same modality, based on textual annotations of each image. Obviously, annotating all parts of the image and complex visual contents is not feasible. As a result, the retrieval performance is limited by the quality and extent of the annotations  \cite{long2003multimedia}. Over the past two decades, major advances in AI, machine vision, and computer hardware have made it possible to apply CBIR in the medical domain with the possibility of search among a huge number of images. Nevertheless, CBMIR is different based on the sensitive semantic gap, the large amount of data, and the lack of labeled datasets  \cite{wei2006content}. 
 
 One of the main steps to overcome the challenges can be addressed by appropriate feature extraction. Features can be divided into two main categories: low-level and high-level \cite{ma2017new}. In the low-level feature extraction category, patch-based methods can be divided into the whole image descriptors (Local Mesh Patterns and Local Binary Patterns, LBP) \cite{murala2014local}) and keypoint-based descriptors such as SIFT, SURF and ORB \cite{kashif2016feature}. However, in the medical domain, feature detection can easily fail to provide keypoints with acceptable quantity, quality, and distribution  \cite{sargent2009feature}. To overcome this problem, dense sampling may be applied. For example dense-SIFT and dense-SURF (applying path descriptors in the whole image) \cite{LI2017100} have been implemented in the medical field frequently.
 
 High-level features, on the other hand, can provide more accurate results due to utilizing the labeled data and reducing the semantic gap. In many works, learning methods have been applied to the low-level features to map or modify them in CBMIR  \cite{rahman2009medical}. Also using the information in the dense layers of deep network architectures has recently become quite popular in CBMIR. The idea is that if a network can achieve good results in classification, then it should be able to extract all necessary features like corners, edges and any other necessary element in every single layer to provide useful information for the last layer  \cite{donahue2014decaf}. Surprisingly, dense layers of pre-trained networks such as VGG and AlexNet which have been trained using very large datasets like ImageNet, can also provide discriminating features and perform well in CBMIR. They have the ability to extract useful information from any input image \cite{kieffer2017convolutional}. As a matter of fact, the quality of features extracted from a pre-trained network might increase by retraining the network, using new labeled data in the applied domain  \cite{kalra2018content}.

\section{Methodology}
This section describes Radon transform as well as the fast and effective method we propose to calculate local Radon projections based on convolution kernels. We also explain the steps and methods to assemble the histogram based on these local Radon projections.
\subsection{Radon Transform}
\begin{figure}[ht]
    \centering
    \includegraphics[width=1\textwidth]{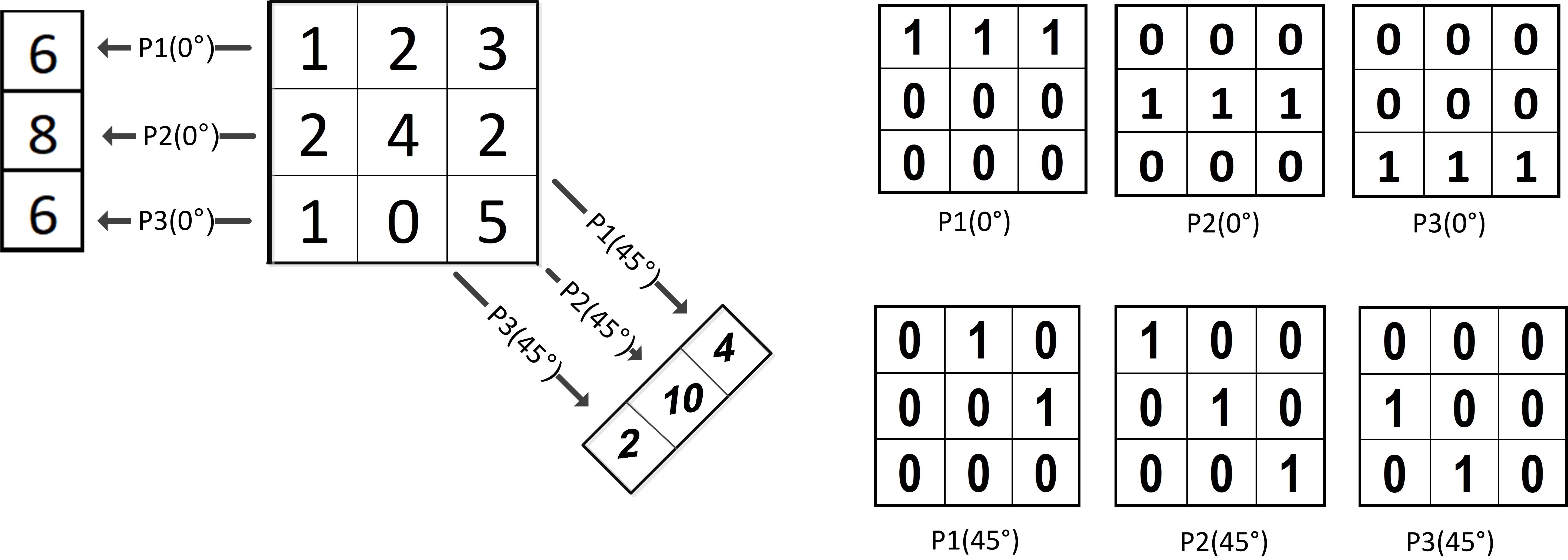}
    \caption{Left: Two estimated Radon projections of a $3\times3$ sample window  at $0^\circ$ and $45^\circ$. Right (Up/Down): Filters to produce $0^\circ$ / $45^\circ$ Radon for each $3 \times 3$ local windows.}
    \label{fig:First}
\end{figure}

In general, Radon transform is described by integrating the projection values of a scene, an image or body parts from various directions \cite{radon1986determination}. The inverse Radon transform along with filtered back-projection has been used to reconstruct the image of internal body parts from projections which were captured at different directions \cite{beylkin1985imaging}. Applying the inverse Radon on values received from an X-ray machine or other modalities is a well-established field in medical imaging. Given the spatial intensities $f(x,y)$, the Radon transform can be formulated as 
\begin{equation}
\mathbf{R}(\rho,\theta)=\int\limits_{-\infty}^{+\infty}\int\limits_{-\infty}^{+\infty} f(x,y)\delta(\rho-xcos\theta-ysin\theta)dxdy,
\label{Radon}
\end{equation}
where $\delta(\cdot)$ is the Dirac delta function. In Figure \ref{fig:First}, the Radon transform is illustrated for a small window.

\subsection{Convolutional Local Radon Projections}
Recently, there has been renewed interest in using Radon transform as an image descriptor \cite{babaieicpram,tizhoosh2015barcode}. Most methods apply Radon transform globally on the entire image or on a relatively large patches \cite{babaie2017local}. However, calculating the Radon transform in quite small local windows would be of interest. We used the local Radon in $3 \times 3$ windows, which resulted in a fast and innovative convolution-based method that calculated Radon in small neighbors. For each direction, we needed a summation of pixels along that direction. For instance, if we wanted to calculate the zero-degree Radon for each pixel of a given image in  $3 \times 3$ windows, we added all $3$ rows in $3 \times 3$ windows (specified by dashed arrows in Figure \ref{fig:First}).

As depicted in Figure \ref{fig:First}, each kernel could easily add one row. Convolving these $3$ kernels, resulting is $3$ images with same size gives us $3$ digits for each pixel which is equivalent to zero-degree Radon. We designed other kernels for $45^{\circ}$, $90^{\circ}$ and $135^{\circ}$ as well. Because there is just one pixel in the first and last element of the Radon projections in  $45^{\circ}$ and $135^{\circ}$, we ignored them to get the same length vector for all $4$ directions (the remaining elements are marked by arrows in Figure \ref{fig:First} for $45^{\circ}$).

\subsection{LRP -- Local Radon Patterns}
\begin{figure*}
	\centering
	\includegraphics[width=1\textwidth]{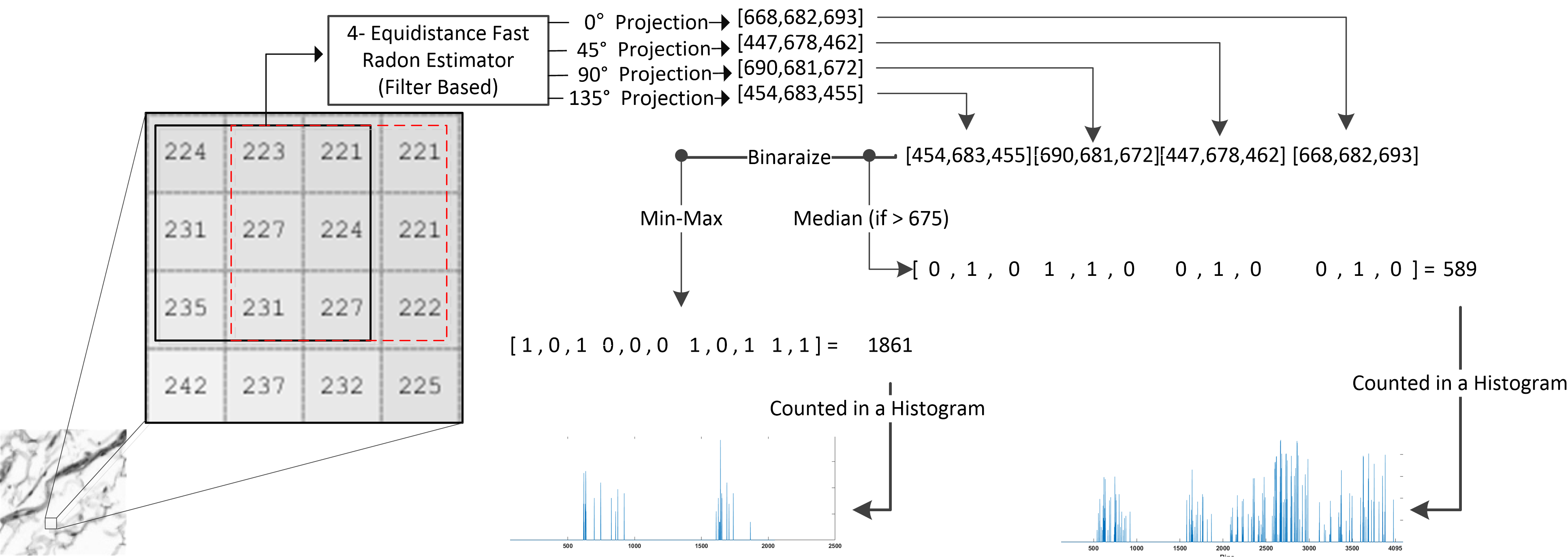}
	\caption{Visualization of the proposed method: Local Radon projections in all directions are calculated by  corresponding kernels.}
	\label{fig:Second}
\end{figure*}

Figure \ref{fig:Second} depicts our proposed method to create the LRP descriptor. Each given image was convolved by three sets of $3 \times 3 $ kernels to obtain the local Radon projections in each direction. Resulting in $12$ kernels were applied for $4$ equidistant directions. Each kernel outcome was an image with the same size as the input image. We concatenated these projections together as explained in Figure  \ref{fig:Second}. 

In the next step, we binarized these 12 numbers to count their equivalent integers in a histogram. We tried two different methods to create a binary vector: Median thresholding and the Min-Max method  \cite{tizhoosh2016minmax}.
The median method is binarizing 12 numbers based on thresholding the vector by its median value. Let's suppose $X=[x_1,x_2,\dots,x_n]$,  then the median method can be formulated as

\[
    B(i)= 
\begin{cases}
    1 ,& \text{if } x(i)\geq \textrm{median}(X)\\
    0,              & \text{otherwise}
\end{cases}
\]

On the other hand, the Min-Max method tries to capture the signal's-concatenated Radon projections- shape, by assigning one if the next adjunct element is greater than the current element and zero if otherwise:
\[
    B(i)= 
\begin{cases}
    1 ,& \text{if } x(i)\geq x(i+1)\\
    0,              & \text{otherwise}
\end{cases}
\]

\section{Experiments and Results}
In order to evaluate our descriptor performance, we selected two publicly available medical datasets. These datasets have been used in related publications. 

\subsection{KimiaPath24 Dataset}
The KimiaPath24 dataset consists patches of 24 whole slides images (WSIs) in pathology domain representing diverse body parts with different texture and stains. The images were captured by TissueScope LE 1.0\footnote{http://www.hurondigitalpathology.com/tissuescope-le-3/} 
bright field using a 0.75 NA lens. For each image, one can determine the resolution by checking the description tag in the header of the file. For instance, if the resolution is 0.5$\mu$m, then the magnification is 20x. The dataset offers 27,055 training patches and 1,325 (manually selected) test patches of size 1000$ \times $1000 ($0.5\mu m \times 0.5\mu m$) \cite{Babaie_2017_CVPR_Workshops}. The  test patches have been removed (whitened) in WSIs such that they cannot be mistakenly used for training. The color (\emph{staining}) is neglected; all patches are saved as grayscale images.  The dataset has a total of $n_\textrm{tot}=1,325$ patches $P_{s}^j$ that belong to 24 sets $\Gamma_s=\{P_{s}^i | s\in S, i=1,2\dots,n_{\Gamma_s}\}$ with $s=0,1,2,\dots,23$. Looking at the set of retrieved images $R$ for any experiment, the \textbf{patch-to-scan accuracy} $\eta_\textrm{p}$ can be given as

\begin{equation}
\label{eqn:etap}
\eta_\textrm{p}=\frac{1}{n_{tot}} \sum\limits_{s\in S} |R \cap \Gamma_s|.
\end{equation}

As well, we calculate the \textbf{whole-scan accuracy} $\eta_\textrm{W}$ as

\begin{equation}
\label{eqn:etaw}
\eta_\textrm{W}=\frac{1}{24} \sum\limits_{s\in S}\frac{|R \cap \Gamma_s|}{n_{\Gamma_s}}.
\end{equation}

\begin{figure}[tb]
\centering 
\stackunder[5pt]{\includegraphics[width=25mm,height=25mm]{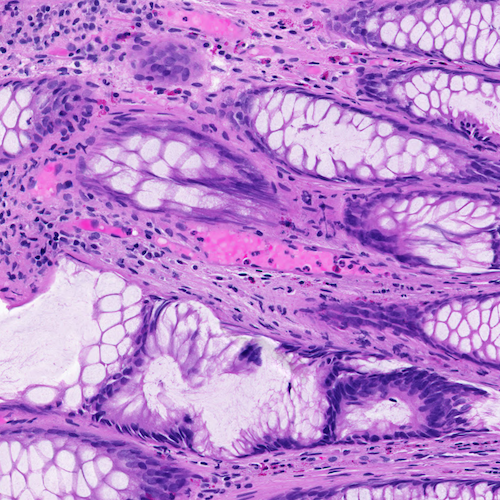}}{\tiny Scan 13}
\stackunder[5pt]{\includegraphics[width=25mm,height=25mm]{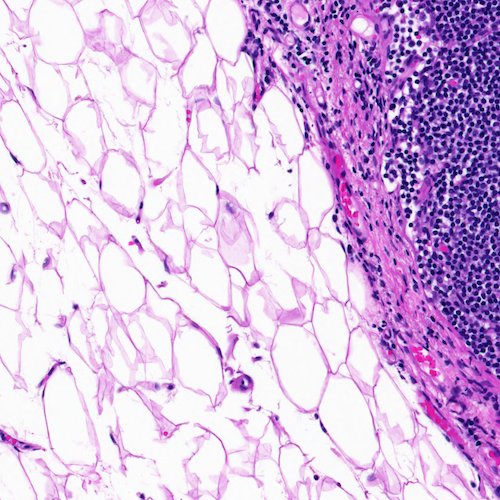}}{\tiny Scan 4}
\stackunder[5pt]{\includegraphics[width=25mm,height=25mm]{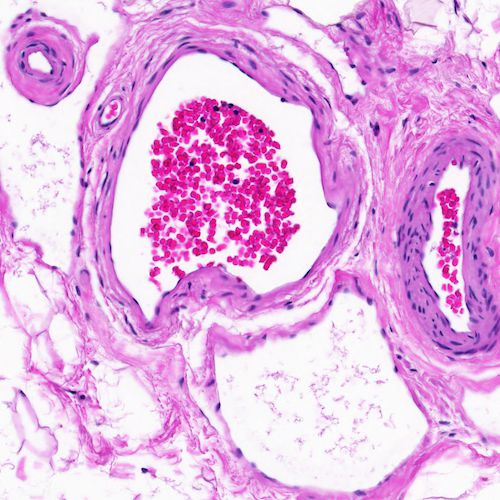}}{\tiny Scan 19} \\ \vspace{0.05in}
\stackunder[5pt]{\includegraphics[width=25mm,height=25mm]{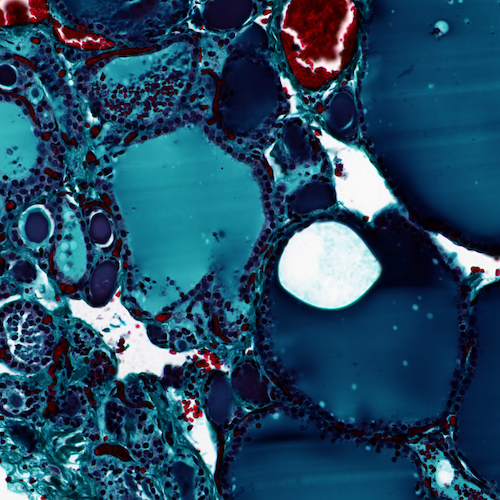}}{\tiny Scan 7}
\stackunder[5pt]{\includegraphics[width=25mm,height=25mm]{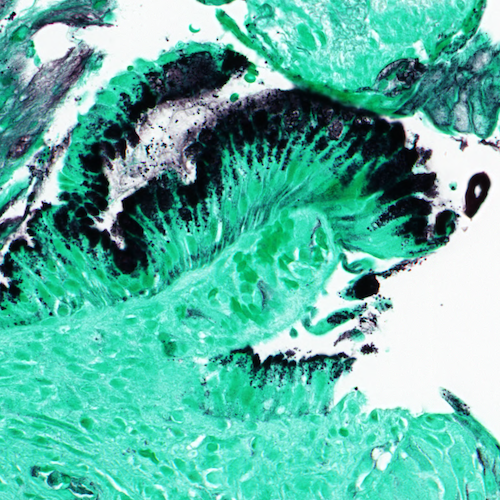}}{\tiny Scan 10}
\stackunder[5pt]{\includegraphics[width=25mm,height=25mm]{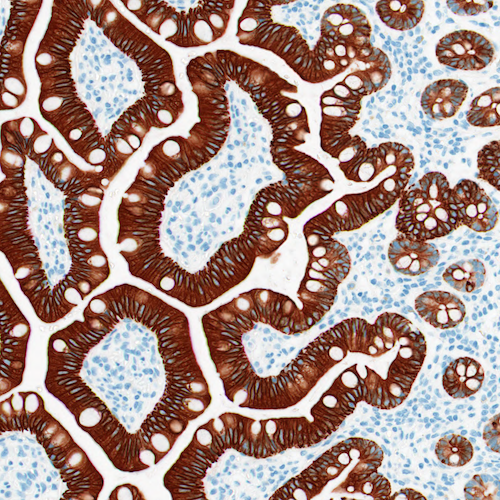}}{\tiny Scan 16}\\
\caption{Sample images from KimiaPath24 dataset.}
\label{fig:SamplePATH24Images}
\end{figure}

Hence, the total accuracy $\eta_\textrm{total}$ (patch-to-scan and whole-scan accuracy)can be defined as  $\eta_\textrm{total}= \eta_\textrm{p}\times\eta_\textrm{W}$.
The dataset and the code for accuracy calculations can be downloaded from the web\footnote{{http://kimia.uwaterloo.ca/}}. Figure \ref{fig:SamplePATH24Images} shows sample patches from KimiaPath24 dataset. We resized images to $250\times 250$ for all methods (for deep network slightly smaller).
\subsection{CT Emphysema Dataset}
\begin{figure}[t]
\centering 
\stackunder[5pt]{\includegraphics[width=25mm,height=25mm]{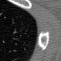}}{\tiny NT}
\stackunder[5pt]{\includegraphics[width=25mm,height=25mm]{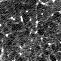}}{\tiny CLE}
\stackunder[5pt]{\includegraphics[width=25mm,height=25mm]{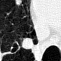}}{\tiny PSE} \\ \vspace{0.05in}
\stackunder[5pt]{\includegraphics[width=25mm,height=25mm]{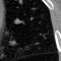}}{\tiny NT}
\stackunder[5pt]{\includegraphics[width=25mm,height=25mm]{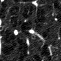}}{\tiny CLE}
\stackunder[5pt]{\includegraphics[width=25mm,height=25mm]{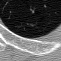}}{\tiny PSE}\\
	\caption{Sample images from CT Emphysema data-set.}
	\label{fig:SampleCTImages}
\end{figure}

Computed tomography emphysema database was introduced by Sorensen et al. \cite{sorensen2010quantitative} to classify lung CT images. A part of this database includes 168 square patches that have been manually annotated in a subset of the slices with an in-plane resolution of 0.78$\times$0.78 mm$^2$, slice thickness of 1.25 mm, tube voltage equal 140 kV, and a tube current of 200 mAs. The 512$\times$512 pixel slices depict the upper, middle, and lower part of the lung of each patient. The 168 patches, of size 61$\times$61 pixels, are from three different classes, NT (normal tissue, 59 observations), CLE (centrilobular emphysema, 50 observations), and PSE (paraseptal emphysema, 59 observations). The NT patches were annotated in \emph{never smokers}, and the CLE and PSE ROIs were annotated in \emph{healthy smokers} and \emph{smokers with COPD} (chronic obstructive pulmonary disease) in areas of the leading pattern. Figure  \ref{fig:SampleCTImages} shows examples for NT, CLE and PSE classes from the CT Emphysema dataset. Given the set of correctly classified images $\mathcal{C}$, the accuracy $A_\textrm{CT}$ can be calculated as 
\begin{equation}
A_\textrm{CT}=\frac{|\mathcal{C}|}{168}.
\end{equation}

\begin{table}[!h!t] 
	\caption{The best results for all datasets and descriptors when city block ($L_1$), Euclidean ($L_2$), Chi-squared ($\chi^2$) and cosine ($\cos$) distances are used for direct similarity measurements. For LBP and HOG, the best results were achieved for each dataset via exhaustive parameter search.}
	\begin{center}
		\begin{tabular}{lcll}
			\toprule
			\multicolumn{3}{l}{\cellcolor{gray!25} \textbf{CT Emphysema Dataset}}\\
			\multicolumn{3}{l}{\cellcolor{gray!25} }\\
			Method & $A_\textrm{CT}$ & $|\mathbf{h}|$ \\	
			\toprule
			\vspace{0.05in}
			LRP(Our method)$_\textrm{MinMax}$ 	 	  	& 82.14\% & 2048	&   \\ \vspace{0.05in}	
			LRP(Our method)$_\textrm{median}$ 	 	  	& 81.32\% & 4096	&   \\ \vspace{0.05in}	
			ELP \cite{tizhoosh2018representing}$_\textrm{median}$ 	 	  	& 80.95\% & 256	&   \\ \vspace{0.05in}
			LBP$^{uri}_{(12,3), \chi^2}$		 	 & 80.36\%	& 18 	&  	\\ \vspace{0.05in}
			VGG-Deep$_{L_2}$\cite{kumar2018deep}			  	& 	69.64\% & 4096	&  \\ 
			\vspace{0.05in}	
			HOG$_{L_2}$			 	  	& 65.47\%	& 1215	&  	\\ 
			\bottomrule
			\multicolumn{3}{l}{\cellcolor{gray!25} \textbf{Kimia Path24 Dataset}}\\ 
			\multicolumn{3}{l}{\cellcolor{gray!25}}\\ 
			Method & $ \{\eta_p,\eta_W\}$ & $|\mathbf{h}|$ &   \\	
			\toprule
			\vspace{0.05in}
			LRP(Our method)$_\textrm{median}$  		  	& \{75.38\%, 72.02\%\} & 4096	 \\ \vspace{0.05in}
			LRP(Our method)$_\textrm{MinMax}$  		  	& \{74.12\%, 73.43\%\} & 2048	 \\ \vspace{0.05in}
			ELP \cite{tizhoosh2018representing}$_\textrm{median}$ 	 	  	& \{71.16\%, 68.05\%\} & 1024	&   \\
			VGG-Deep$_{\cos}$\cite{kumar2018deep}			  	& \{70.11\%, 68.13\%\}	 & 4096	&  	\\ \vspace{0.05in}	
			LBP$^u_{(24,2), L_1}$		 	 & \{65.55\%, 62.56\%\}	 & 555	&  	\\ \vspace{0.05in}	
			HOG$_{L_1}$			 	  	& \{17.58\%, 16.76\%\}	 & 648	&  	\\ 
			
		\end{tabular}
	\end{center}
	\label{tab:subimages}
\end{table}%

\subsection{Results}
Table \ref{tab:subimages} represents our experimental results. We performed image search and selected the label of the best match image (top-1 accuracy). For the CT Emphysema dataset the leave-one-out strategy was chosen. While for the KimaPath24 test images are searched on the training set. 
The LRP method does not have any parameter to tune except the method of binarization. On the other hand, for LBP and HoG presented results, we did exhaustive search to tune their parameters. As Table \ref{tab:subimages} suggests, both binarization methods provided  superior results in comparison to other methods. Moreover, LRP  benefits from fast kernel-based calculation time. LRP takes 0.01 seconds for a $1000 \times 1000$ while the optimized ELP (encoded local projections) takes 90 seconds for the same image size mainly due to the Radon computation time. 

\section{Conclusions}
In this work, we introduced a fast and simple histogram-based descriptor based on the local Radon transform and applied it in two medical datasets with different modalities. In the im
developed local Radon descriptor) in a much shorter time. 

\bibliographystyle{splncs04} 
\bibliography{reference.bib}

\end{document}